\pdfoutput=1
\documentclass[a4paper]{article}

\usepackage{INTERSPEECH2022}
\usepackage{tikz}
\usetikzlibrary{arrows.meta,positioning,fit,backgrounds,calc,decorations.pathreplacing}

\usepackage{xcolor}
\usepackage{url}
\usepackage{amssymb}
\usepackage[most]{tcolorbox}
\usepackage{tabularx}

\title{SEAR: Segment-Evidence-Aware Routing for Weak-to-Strong
       Multilingual Speech MCQ}

\name{Huy Hoang Le$^{1,*}$, Long-Bao Nguyen$^{1,*}$, Minh Tri Dao$^{1}$}

\address{
  $^1$CAKE by VPBank, Vietnam\\
  $^*$Equal contribution
}

\email{hoang.le2@cake.vn, bao.nguyen@cake.vn, tri.dao@cake.vn}

\begin{document}

\maketitle

\begin{abstract}
This paper describes our system for Task~2 of the second Multilingual
Conversational Speech Language Model (MLC-SLM) Challenge. We adapt
Qwen3-Omni-30B-A3B-Instruct with a segment-evidence-aware data and post-training
pipeline. A language model converts timestamped ASR into coherent event spans,
which are expanded by a boundary margin and cropped from the original recording.
We then synthesize complementary semantic MCQs with Qwen3.6-27B and acoustic MCQs
with Gemini~3.1 Flash-Lite, followed by structural, grounding, answer-consistency,
and target-model trainability checks, yielding 359{,}825 verified MCQs across 21
language and accent variants. A text-only probe partitions the data into weak,
text-answerable items used for supervised fine-tuning and strong, audio-dependent
items used for reinforcement learning with Group Sequence Policy Optimization
(GSPO), stabilized by debiased advantages, sequence-level importance correction,
and dynamic filtering.
Our system obtains 90.92\% accuracy on the final official evaluation set.
\end{abstract}

\section{Introduction}

\begin{figure*}[!t]
  \centering
  \resizebox{\textwidth}{!}{%
  \begin{tikzpicture}[
    font=\footnotesize,
    >={Stealth[length=2mm]},
    node distance=3.5mm and 7mm,
    data/.style   ={rectangle, rounded corners=2pt, draw=blue!45, fill=blue!6,
                    text width=18mm, align=center, minimum height=11mm, inner sep=2.5pt},
    mod/.style    ={rectangle, rounded corners=3pt, draw=violet!55, fill=violet!8,
                    text width=18mm, align=center, minimum height=11mm, thick, inner sep=2.5pt},
    weak/.style   ={rectangle, rounded corners=2pt, draw=orange!65!black, fill=orange!12,
                    text width=19mm, align=center, minimum height=10mm, inner sep=2.5pt},
    strong/.style ={rectangle, rounded corners=2pt, draw=teal!70!black, fill=teal!10,
                    text width=19mm, align=center, minimum height=10mm, inner sep=2.5pt},
    arr/.style    ={->, thick, gray!65},
    elab/.style   ={font=\scriptsize\itshape, text=gray!55!black, inner sep=1.5pt},
    phasebox/.style={rounded corners=4pt, draw=gray!55, dashed, fill=gray!4, inner sep=2.5mm},
    phaselab/.style={font=\scriptsize\bfseries, text=gray!45!black, inner sep=1pt}
  ]
    \node[data] (conv) {Raw audio\\ {\scriptsize+ timestamped ASR}};
    \node[mod, right=of conv] (seg) {LLM event spans\\ {\scriptsize+ boundary margin}};
    \node[mod, right=9mm of seg] (gen) {Dual-branch MCQ\\ synthesis\\ {\scriptsize semantic / acoustic}};
    \node[mod, right=of gen] (ver) {Quality \&\\ trainability\\ verification};
    \node[data, right=of ver] (mcq) {Verified segment\\ Audio MCQ\\ {\scriptsize$\langle q,o,a\rangle$}};
    \node[mod, right=9mm of mcq] (probe) {Text-only\\ probe\\ {\scriptsize(audio-free)}};
    \node[weak,   above right=3mm and 8mm of probe] (weak)   {\textbf{weak}\\ {\scriptsize text-answerable}\\ $\rightarrow$ SFT};
    \node[strong, below right=3mm and 8mm of probe] (strong) {\textbf{strong}\\ {\scriptsize audio-dependent}\\ $\rightarrow$ GSPO};

    \draw[arr] (conv) -- (seg);
    \draw[arr] (seg)  -- (gen) node[midway, above, elab] {$\{s_i\}$};
    \draw[arr] (gen)  -- (ver);
    \draw[arr] (ver)  -- (mcq);
    \draw[arr] (mcq)  -- (probe);
    \draw[arr] (probe.east) -- (weak.west)   node[midway, above, elab] {solved};
    \draw[arr] (probe.east) -- (strong.west) node[midway, below, elab] {failed};

    \begin{scope}[on background layer]
      \node[phasebox, fit=(conv)(seg)(conv.west |- weak.north)(seg.east |- strong.south)] (p1) {};
      \node[phasebox, fit=(gen)(mcq)(gen.west |- weak.north)(mcq.east |- strong.south)]   (p2) {};
      \node[phasebox, fit=(probe)(weak)(strong)] (p3) {};
    \end{scope}
    \node[phaselab, anchor=south west] at (p1.north west) {Stage 1 $\cdot$ Event Localization};
    \node[phaselab, anchor=south west] at (p2.north west) {Stage 2 $\cdot$ Dual-Branch Synthesis \& Verification};
    \node[phaselab, anchor=south west] at (p3.north west) {Stage 3 $\cdot$ Audio-Contribution Routing};
  \end{tikzpicture}%
  }
  \caption{\textbf{Segment-level Audio MCQ curation and routing.}
  \textbf{Stage~1:} timestamped ASR is used by an LLM to identify coherent
  conversational events; each event is expanded by a boundary margin before the
  waveform is cropped, and model-facing timestamps are rebased to the crop.
  \textbf{Stage~2:} Qwen3.6-27B generates semantic MCQs from the event transcript
  and metadata, while Gemini~3.1 Flash-Lite generates acoustic MCQs from the audio
  segment. Candidates pass grounding, uniqueness, distractor, format, timestamp,
  and target-model trainability checks. \textbf{Stage~3:} an audio-free probe
  estimates audio contribution. Weak items are used for SFT; strong items are
  candidates for GSPO, with online dynamic filtering retaining only rollout groups
  with non-zero reward variance.}
  \label{fig:pipeline}
\end{figure*}
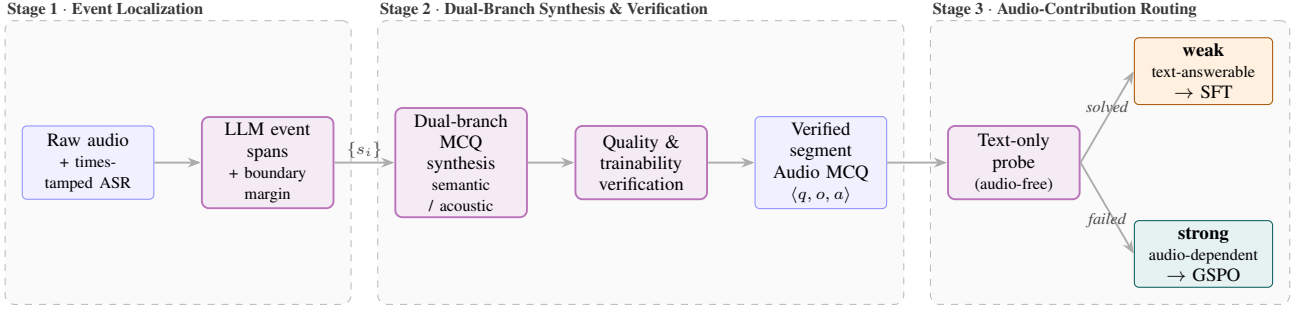

Spoken language understanding over conversational speech requires a system to
jointly interpret acoustic content and linguistic meaning. Task~2 of the MLC-SLM
Challenge instantiates this problem as multiple-choice question answering: given a
multilingual two-speaker conversation and a question about it, a system must select
the correct option, and performance is measured by answer accuracy~\cite{mlcslm2026}.
The task is demanding because the conversations span fourteen languages, are informal
and spontaneous, and the questions probe both \emph{what} was said and \emph{how} it
was said, so a system cannot rely on transcription alone.

The central difficulty for audio LLMs on this task is that strong text priors let a
model answer many questions without genuinely listening. Recent analysis shows that
large audio language models frequently produce correct answers from textual cues
while ignoring the audio, a ``zero audio-contribution'' effect that inflates apparent
accuracy yet fails on acoustically grounded questions~\cite{he2025audiomcq}. Treating
every training example identically therefore wastes supervision on questions the model
could already answer from text, and dilutes the signal from the questions that truly
require listening.

We address this with Qwen3-Omni-30B-A3B-Instruct~\cite{qwen2025qwen3omni}
and a segment-evidence-aware post-training pipeline. Timestamped ASR is converted
into coherent event spans, expanded with boundary context, and cropped into
training clips. Semantic questions are generated from transcript evidence, whereas
acoustic questions are generated directly from the waveform. A multi-stage verifier
rejects ungrounded, ambiguous, malformed, or trivially leaked questions and measures
whether the target model can produce informative mixed-reward rollouts. We then use
supervised fine-tuning (SFT) for instruction and output-format adaptation, followed
by GSPO-based reinforcement learning on audio-dependent questions.

Our contributions are as follows.
\textbf{(i)}~A timestamp-aware, event-preserving segmentation pipeline that uses
LLM-extracted event spans and boundary margins instead of arbitrary fixed windows.
\textbf{(ii)}~A dual-branch synthesis and verification engine: Qwen3.6-27B~\cite{qwen2026qwen36} creates semantic MCQs,
Gemini~3.1 Flash-Lite~\cite{google2026gemini31flashlite} creates acoustic MCQs, and target-aware verification
separates valid-but-easy items from trainable RL items.
\textbf{(iii)}~A stable weak-to-strong post-training recipe of SFT followed by
GSPO-based reinforcement learning. The resulting system reaches 90.92\%
accuracy on the final official evaluation set.

\section{Related Work}
\textbf{Audio-language data construction.}
AudioMCQ converts captioned audio into four-option questions and filters candidates
for answer consistency, distractor quality, fluency, and reasoning fidelity
~\cite{he2025audiomcq}. Our pipeline extends data construction to long multilingual
conversations: timestamped transcripts define coherent candidate event spans, while
separate text and audio generators specialize in semantic and acoustic questions.

\textbf{Audio-contribution-aware post-training.}
AudioMCQ identifies zero audio-contribution cases by testing whether a question can
be answered without the original audio and proposes weak-to-strong and
mixed-to-strong schedules~\cite{he2025audiomcq}. We retain this audio-contribution
axis, but add a separate target-model trainability test because audio dependence does
not guarantee useful group-relative reward variance.

\textbf{Sequence-level and debiased group-relative reinforcement learning.}
GRPO normalizes rewards within a response group and uses token-level policy ratios.
GSPO replaces these with a length-normalized sequence likelihood ratio and clips the
whole response, improving stability for MoE models~\cite{zheng2025gspo,qwen2025gspoblog}.
Dr.GRPO removes the response-length and within-group standard-deviation
normalizations that bias vanilla GRPO~\cite{liu2025drgrpo}, and sequence-level
truncated importance sampling (TIS) corrects the numerical mismatch between the
rollout engine and the training backend~\cite{feng2025offpolicy}. Our recipe
combines these three corrections (Section~\ref{sec:curriculum}).

\section{Method}

\subsection{Base model and parameter-efficient adaptation}
\label{sec:backbone}
We adopt Qwen3-Omni-30B-A3B-Instruct, a Mixture-of-Experts audio LLM with roughly
30B total and 3B activated parameters, whose state-of-the-art multilingual audio
understanding matches the fourteen-language scope of the
challenge~\cite{qwen2025qwen3omni}. We adapt only the language model with low-rank
adaptation (LoRA), keeping the pretrained audio encoder and multimodal aligner
frozen: fine-tuning them on the comparatively small challenge set risks
catastrophic forgetting of large-scale multilingual acoustic knowledge.

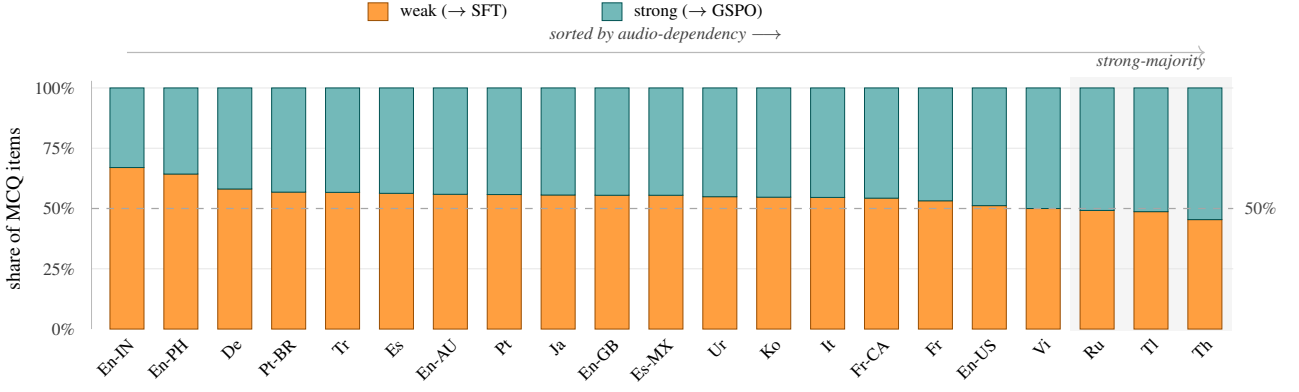
\begin{figure*}[!t]
  \centering
  \resizebox{\textwidth}{!}{%
  \begin{tikzpicture}[
    font=\footnotesize,
    wk/.style={fill=orange!75, draw=orange!60!black, line width=.3pt},
    st/.style={fill=teal!55,  draw=teal!65!black,  line width=.3pt},
  ]
    \fill[gray!8] (13.30,-0.03) rectangle (15.58,3.55);
    \node[font=\scriptsize\itshape, text=gray!55!black, anchor=south]
      at (14.44,3.55) {strong-majority};
    \foreach \p in {25,75,100}{
      \draw[gray!20, line width=.3pt] (-0.5,\p*0.034) -- (15.6,\p*0.034);}
    \draw[gray!55] (-0.5,0) -- (-0.5,3.5);
    \foreach \p in {0,25,50,75,100}{
      \node[font=\scriptsize, text=gray!45!black, anchor=east] at (-0.62,\p*0.034) {\p\%};}
    \node[font=\footnotesize, rotate=90, anchor=south] at (-1.35,1.7) {share of MCQ items};
    \foreach \lab/\wk [count=\i from 0] in {%
      En-IN/67.0, En-PH/64.3, De/58.1, Pt-BR/56.8, Tr/56.7, Es/56.3, En-AU/55.9,
      Pt/55.8, Ja/55.6, En-GB/55.5, Es-MX/55.5, Ur/54.9, Ko/54.7, It/54.6,
      Fr-CA/54.3, Fr/53.2, En-US/51.2, Vi/50.0, Ru/49.2, Tl/48.7, Th/45.4}{
        \pgfmathsetmacro{\xc}{\i*0.76}
        \pgfmathsetmacro{\hw}{\wk*0.034}
        \draw[wk] (\xc-0.24,0)   rectangle (\xc+0.24,\hw);
        \draw[st] (\xc-0.24,\hw) rectangle (\xc+0.24,3.4);
        \node[font=\scriptsize, rotate=45, anchor=north east, inner sep=1pt]
          at (\xc,-0.06) {\lab};}
    \draw[gray!70, dashed, line width=.5pt] (-0.5,1.7) -- (15.6,1.7);
    \node[font=\scriptsize, text=gray!55!black, anchor=west] at (15.65,1.7) {50\%};
    \draw[->, gray!55, line width=.5pt] (0,3.9) -- (15.2,3.9);
    \node[font=\scriptsize\itshape, text=gray!50!black, anchor=south] at (7.6,3.92)
      {sorted by audio-dependency $\longrightarrow$};
    \draw[wk] (3.4,4.35) rectangle (3.68,4.6);
    \node[anchor=west, font=\scriptsize] at (3.74,4.475) {weak ($\rightarrow$ SFT)};
    \draw[st] (6.7,4.35) rectangle (6.98,4.6);
    \node[anchor=west, font=\scriptsize] at (7.04,4.475) {strong ($\rightarrow$ GSPO)};
  \end{tikzpicture}%
  }
  \caption{\textbf{Audio-dependency generalizes across all 21 languages.}
  Each bar is one language/accent variant, normalized to 100\% and sorted by
  \emph{weak} (text-answerable, orange $\rightarrow$ SFT) share; the remainder is
  \emph{strong} (audio-dependent, teal $\rightarrow$ GSPO). The weak fraction exceeds the
  strong fraction for 18 of 21 variants; only Thai, Russian, and Tagalog are
  strong-majority (shaded). The full corpus contains 359{,}825 segment-level Audio MCQ
  items, balanced across languages (11.5K--21.9K per variant). Codes: En-IN{\,=\,}Indian English,
  En-PH{\,=\,}Filipino English, Tl{\,=\,}Tagalog, etc.}
  \label{fig:dataset}
\end{figure*}

\subsection{Segment-level Audio MCQ curation}
\label{sec:data}
Our data engine uses timestamp-guided, event-preserving segmentation for long
conversational speech. It produces 359{,}825 verified segment-level MCQs across
21 language and accent variants, with complementary semantic and acoustic question
families.

\textbf{Timestamped ASR and event localization.}
For each raw conversation, an ASR system produces timestamped transcript units
$\{(u_j,t_j^{\mathrm{s}},t_j^{\mathrm{e}})\}$. An LLM reads the ordered transcript
and returns self-contained conversational events as intervals
$e_k=[b_k^{\mathrm{s}},b_k^{\mathrm{e}}]$. We expand each interval by a per-event
context margin $m_k$, varied rather than fixed so that cropped segments span
diverse durations, and clip it to the recording bounds,
\begin{equation}
  s_k=\bigl[\max(0,b_k^{\mathrm{s}}-m_k),
             \min(T,b_k^{\mathrm{e}}+m_k)\bigr],
  \label{eq:segment}
\end{equation}
then merge strongly overlapping windows. This preserves complete turns and nearby
prosodic context while avoiding unrelated portions of the conversation. Every
model-facing timestamp is rebased to the crop as
$t^{\mathrm{local}}=t^{\mathrm{abs}}-s_k^{\mathrm{s}}$ (the absolute offset is kept
only as provenance), so timestamps always refer to the beginning of the waveform
actually supplied to the model, matching evaluation conditions.

\textbf{Dual-branch MCQ synthesis.}
The semantic branch supplies Qwen3.6-27B~\cite{qwen2026qwen36} with the localized transcript, event span,
and conversation context. It generates four-option questions about content, intent,
facts, discourse relations, and temporal order. The acoustic branch supplies the
cropped waveform to Gemini~3.1 Flash-Lite~\cite{google2026gemini31flashlite} and generates questions about speaker identity,
pitch, speaking rate, emotion, voice quality, background sound, overlap, and other
audible events. Each candidate is serialized as
$z=(s,q,\{o_A,o_B,o_C,o_D\},a,\tau)$, where $a$ is the keyed option and $\tau$
contains provenance, absolute and rebased timestamps, language, and question type. Correct-answer
positions are permuted to reduce option-position bias.

\textbf{Quality and trainability verification.}
Verification serves two purposes. First, an LLM judge checks that each candidate is
grounded in the supplied evidence: the question is answerable, exactly one option is
correct, distractors are plausible but unsupported, rebased timestamps fall inside
the clip, the language is fluent, and neither the stem nor the options leak the key.
Malformed or ambiguous candidates are rejected or regenerated. This mirrors the
in-pipeline grading of AudioMCQ, with an additional waveform-grounding check for
acoustic questions~\cite{he2025audiomcq}.

Second, we probe \emph{trainability} with the target Qwen3-Omni-30B-A3B-Instruct model using $G=8$ audio-conditioned responses per candidate, labeling each as solved (all correct), learnable (mixed), or unsolved (all wrong). Solved items remain in SFT but are excluded from the initial GSPO queue to reduce rollout cost. Learnable items form the primary GSPO pool, while unsolved items are retained and dynamically admitted as they become learnable during training (Section~\ref{sec:curriculum}).

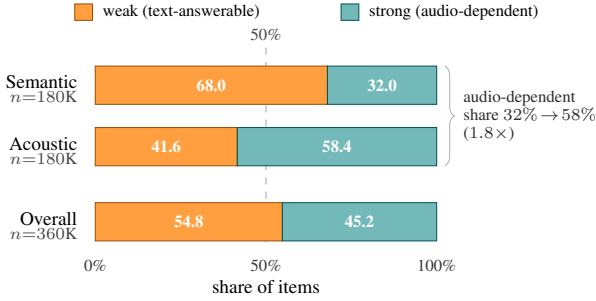
\begin{figure}[t]
  \centering
  \resizebox{\linewidth}{!}{%
  \begin{tikzpicture}[
    font=\footnotesize,
    wk/.style={fill=orange!75, draw=orange!60!black, line width=.3pt},
    st/.style={fill=teal!55,  draw=teal!65!black,  line width=.3pt},
    pin/.style={font=\scriptsize\bfseries, text=white},
  ]
    \def\W{5.0}   %
    \draw[gray!60, dashed, line width=.5pt] (0.5*\W,0.55) -- (0.5*\W,-2.35);
    \node[font=\scriptsize, text=gray!55!black, anchor=south] at (0.5*\W,0.55) {50\%};
    \draw[wk] (0,-0.28)       rectangle (0.680*\W,0.28);
    \draw[st] (0.680*\W,-0.28) rectangle (\W,0.28);
    \node[pin] at (0.340*\W,0) {68.0}; \node[pin] at (0.840*\W,0) {32.0};
    \node[anchor=east, font=\footnotesize] at (-0.15,0.05) {Semantic};
    \node[anchor=east, font=\scriptsize, text=gray!55!black] at (-0.15,-0.17) {$n{=}180$K};
    \draw[wk] (0,-1.18)       rectangle (0.416*\W,-0.62);
    \draw[st] (0.416*\W,-1.18) rectangle (\W,-0.62);
    \node[pin] at (0.208*\W,-0.9) {41.6}; \node[pin] at (0.708*\W,-0.9) {58.4};
    \node[anchor=east, font=\footnotesize] at (-0.15,-0.85) {Acoustic};
    \node[anchor=east, font=\scriptsize, text=gray!55!black] at (-0.15,-1.07) {$n{=}180$K};
    \draw[wk] (0,-2.28)       rectangle (0.548*\W,-1.72);
    \draw[st] (0.548*\W,-2.28) rectangle (\W,-1.72);
    \node[pin] at (0.274*\W,-2.0) {54.8}; \node[pin] at (0.774*\W,-2.0) {45.2};
    \node[anchor=east, font=\footnotesize] at (-0.15,-1.95) {Overall};
    \node[anchor=east, font=\scriptsize, text=gray!55!black] at (-0.15,-2.17) {$n{=}360$K};
    \foreach \p in {0,50,100}{
      \node[font=\scriptsize, text=gray!45!black, anchor=north] at (\p/100*\W,-2.44) {\p\%};}
    \node[font=\footnotesize, anchor=north] at (0.5*\W,-2.74) {share of items};
    \draw[wk] (-0.30,0.95) rectangle (-0.04,1.19);
    \node[anchor=west, font=\scriptsize] at (0.00,1.07) {weak (text-answerable)};
    \draw[st] (3.60,0.95) rectangle (3.86,1.19);
    \node[anchor=west, font=\scriptsize] at (3.90,1.07) {strong (audio-dependent)};
    \draw[decorate, decoration={brace, amplitude=4pt}, gray!65]
      (\W+0.12,0.28) -- (\W+0.12,-1.18);
    \node[anchor=west, align=left, font=\scriptsize, text=gray!35!black]
      at (\W+0.28,-0.45) {audio-dependent\\ share $32\%\!\rightarrow\!58\%$\\ ($1.8\times$)};
  \end{tikzpicture}%
  }
  \caption{\textbf{Question type predicts audio dependence}. The text-only probe labels items as weak (text-answerable) or strong (audio-dependent). Audio-dependent items increase from 32.0\% of semantic questions to 58.4\% of acoustic questions ($1.8\times$), validating the probe and providing GSPO with 162,594 well-defined strong items.}
  \label{fig:semaco}
\end{figure}

\subsection{Audio-contribution routing}
\label{sec:partition}
We estimate whether the audio is necessary by running an audio-free probe on the
question and four options. An item answered correctly without the waveform is labeled
\emph{weak}; otherwise it is labeled \emph{strong}. Across the corpus, 197{,}231
items are weak and 162{,}594 are strong. The weak rate is 68.0\% for semantic
questions but only 41.6\% for acoustic questions (Figure~\ref{fig:semaco}), which
supports the intended distinction between transcript-recoverable and perceptually
grounded supervision.

Audio contribution and trainability are orthogonal. A strong item can still be
all-wrong under the current policy and therefore provide zero group-relative signal;
a weak item can still be useful for SFT. We therefore use the text-only probe for
\emph{stage routing} and the $G=8$ target-model probe only for \emph{compute-aware
selection within a stage}.

\subsection{Supervised fine-tuning}
\label{sec:sft}
The first stage adapts the pretrained model to the challenge instruction, multilingual
question style, and exact answer format; the assistant target is the minimal sequence
\texttt{<answer>X</answer>}, where $X\in\{A,B,C,D\}$. We train on all verified weak
items.

\subsection{Reinforcement learning}
\label{sec:curriculum}
For each strong prompt $x$, we sample $G=8$ rollouts $\{y_i\}_{i=1}^{G}$ from the
old policy, scored with a binary exact-match reward,
\begin{equation}
 r_i=\mathbf{1}\!\left[y_i=\texttt{<answer>}a\texttt{</answer>}\right],
 \label{eq:reward}
\end{equation}
where malformed outputs receive zero. Because this reward attaches to the complete
answer sequence, we optimize with GSPO, whose sequence-level ratio and clipping
stabilize RL for MoE models such as Qwen3-Omni, where small expert-routing changes
make token-level ratios volatile~\cite{zheng2025gspo,qwen2025gspoblog}.

\textbf{Advantage debiasing.}
Vanilla GRPO standardizes each reward by the within-group standard deviation,
$\hat A_i^{\mathrm{GRPO}}=(r_i-\bar r)/(\sigma_r+\eta)$, which for binary rewards
and a small group over-amplifies a single discordant rollout even when it stems from
ambiguity, label noise, or stochastic decoding. Following
Dr.GRPO~\cite{liu2025drgrpo}, we keep only the group-relative baseline,
\begin{equation}
 \hat A_i=r_i-\bar r, \qquad
 \bar r=\frac{1}{G}\sum_{j=1}^{G}r_j,
 \label{eq:advantage}
\end{equation}
so the update magnitude reflects the actual disagreement in the group. Dr.GRPO's
length-normalization correction is immaterial here because valid completions share
the same short format.

\textbf{Sequence-level objective.}
GSPO defines the length-normalized sequence ratio
\begin{equation}
\begin{aligned}
 d_{i,t} &= \log \pi_\theta(y_{i,t}\mid x,y_{i,<t})
          - \log \pi_{\mathrm{old}}(y_{i,t}\mid x,y_{i,<t}),\\
 s_i(\theta) &= \exp\!\left(\frac{1}{|y_i|}\sum_{t=1}^{|y_i|}d_{i,t}\right),
\end{aligned}
 \label{eq:gspo_ratio}
\end{equation}
and clips the entire response rather than individual
tokens~\cite{zheng2025gspo}. Combined with the mean-centered advantage, the objective is
\begin{equation}
\begin{aligned}
 \ell_i &= \min\!\left(s_i\hat A_i,
 \operatorname{clip}(s_i,1-\epsilon,1+\epsilon_{\mathrm{high}})\hat A_i\right),\\
 \mathcal{J} &= \mathbb{E}\!\left[\frac{1}{G}\sum_{i=1}^{G}\ell_i\right].
\end{aligned}
 \label{eq:gspo_obj}
\end{equation}

\textbf{Sequence-level TIS.}
Rollouts are generated by vLLM while gradients are
evaluated by the training backend, so even with synchronized weights, numerical
differences can yield $\pi_{\mathrm{vLLM}}\neq\pi_{\mathrm{train}}$. Following
truncated importance sampling (TIS)~\cite{feng2025offpolicy}, we correct this
mismatch with the capped sequence-level training-to-rollout ratio
\begin{equation}
\begin{aligned}
 q_{i,t} &= \log \pi_{\mathrm{train}}(y_{i,t}\mid x,y_{i,<t})
          - \log \pi_{\mathrm{vLLM}}(y_{i,t}\mid x,y_{i,<t}),\\
 c_i &= \exp\!\left(\frac{1}{|y_i|}\sum_t q_{i,t}\right),
 \qquad \tilde c_i=\min(c_i,2).
\end{aligned}
 \label{eq:tis}
\end{equation}
and multiply each response term in Equation~\ref{eq:gspo_obj} by $\tilde c_i$,
preventing backend-mismatched trajectories from dominating the gradient.

\textbf{Dynamic filtering.}
All-correct and all-wrong groups have $\hat A_i=0$ for every response and produce no
group-relative gradient. At every update we discard groups with
$\sigma_r\leq\delta$ and over-sample replacement prompts until the effective batch
is full, following DAPO's dynamic sampling principle~\cite{yu2025dapo}. Because
policy competence changes during training, this online rule is more reliable than
permanently deleting every item the base model answers correctly.

\section{Experiments}

\subsection{Data and metric}

\textbf{Training data.} We train on the curated segment-level Audio MCQ corpus of
Section~\ref{sec:data}: 359{,}825 items synthesized from the Task~2 training conversations,
spanning 21 language and accent variants and balanced across them (11.5K--21.9K items per
variant; Figure~\ref{fig:dataset}). The text-only probe partitions the corpus into
197{,}231 \emph{weak} (text-answerable) and 162{,}594 \emph{strong} (audio-dependent) items
(54.8\% weak overall; Figure~\ref{fig:semaco}), which feed SFT (Section~\ref{sec:sft}) and GSPO
(Section~\ref{sec:curriculum}).

\textbf{Evaluation.} We evaluate on the official MLC-SLM Task~2 development and evaluation
sets and report accuracy, the official metric~\cite{mlcslm2026}.

\subsection{Implementation details}
Our backbone is Qwen3-Omni-30B-A3B-Instruct~\cite{qwen2025qwen3omni}. We train LoRA
adapters on the language model with rank $32$, $\alpha=64$, and dropout $0.05$,
while freezing the audio encoder and aligner. SFT uses a learning rate of
$1\times10^{-4}$, cosine decay, warmup ratio $0.05$, weight decay $0.01$, gradient
clipping at $1.0$, a maximum sequence length of 32{,}768, and \texttt{bf16}.

For RL, we use GSPO with $G=8$, a rollout batch size of 64,
$\epsilon=3\times10^{-4}$, $\epsilon_{\mathrm{high}}=4\times10^{-4}$, and $\beta=0$. We apply Dr.GRPO-style
unnormalized group advantages, sequence-level rollout
importance correction with TIS threshold 2, and dynamic filtering of groups whose reward
standard deviation is at or near zero, monitoring retained-group rate, reward,
clipping fraction, sequence-level mismatch, and effective sample size throughout
training. The reward parser accepts exactly one option in
\texttt{<answer>A/B/C/D</answer>}. Rollout and training use the same tokenizer,
chat template, stop tokens, and synchronized model weights.

\subsection{Main results}
Table~\ref{tab:main} reports accuracy on the development and evaluation sets. Our
system reaches 90.92\% on the Phase-2 (final) evaluation set and ranks
2nd on the official leaderboard.

\begin{table}[t]
  \caption{Task~2 accuracy (\%). Eval is the Phase-1 evaluation set for the
  baselines and the Phase-2 (final) evaluation set for our system.
  \textbf{Bold} marks the best result.}
  \label{tab:main}
  \centering
  \footnotesize
  \setlength{\tabcolsep}{4pt}
  \begin{tabular}{lcc}
    \toprule
    \textbf{System} & \textbf{Dev}~$\uparrow$ & \textbf{Eval}~$\uparrow$ \\
    \midrule
    Qwen2.5-Omni-7B (baseline)~\cite{xu2025qwen25omni}      & 83.16 & 56.42 \\
    Qwen3-Omni-30B-A3B (zero-shot)~\cite{qwen2025qwen3omni} & 92.40 & 77.36 \\
    \textbf{Ours (SEAR)}                                    & \textbf{96.02} & \textbf{90.92} \\
    \bottomrule
  \end{tabular}
\end{table}

\subsection{Ablation study}
Table~\ref{tab:ablation} reports a compact cumulative ablation covering the two
main decisions in our system: curated segment-level supervision, and the
weak-to-strong RL stage. The latter is reported as a single step---routing plus the
stabilized GSPO objective (debiasing, TIS, dynamic filtering)---rather than as a
combinatorial sweep over its components. We also include an all-data control that trains on the same data without weak/strong classification. It achieves only $88.24$ Eval, $2.68$ points below SEAR ($90.92$), confirming that the classification drives the improvement.

\begin{table}[t]
  \caption{Cumulative ablation. The middle row is an all-data control that omits
  the weak/strong classification. $\Delta$ is against the preceding row; the
  zero-shot and curated-SFT rows use the Phase-1 evaluation set, the control and
  final row the Phase-2 (final) set.}
  \label{tab:ablation}
  \centering
  \footnotesize
  \setlength{\tabcolsep}{4pt}
  \begin{tabular}{lcc}
    \toprule
    \textbf{Configuration} & \textbf{Eval}~$\uparrow$ & \textbf{$\Delta$} \\
    \midrule
    Qwen3-Omni-30B-A3B, zero-shot                    & 77.36 & -- \\
    \midrule
    + SFT full data, no weak-strong split          & \underline{88.24} & +10.88 \\
    \midrule
    + verified event-segment MCQ SFT                 & 86.55 & +9.19 \\
    + weak-to-strong GSPO RL (SEAR)                  & \textbf{90.92} & +4.37 \\
    \bottomrule
  \end{tabular}
\end{table}

\subsection{Model training prompts}
\label{sec:prompts}
We use the same answer-only template, independent of language and question type,
for SFT and GSPO rollouts; only SFT includes the gold assistant response. This
prevents prompt-format differences from confounding the training-stage ablation.

\begin{tcolorbox}[
    colback=gray!3,
    colframe=black!40,
    boxrule=0.6pt,
    arc=1.5mm,
    left=1.5mm,
    right=1.5mm,
    top=1mm,
    bottom=1mm,
    title=\textbf{Audio MCQ Training Prompt},
    fonttitle=\small,
    fontupper=\small
]
\begin{tabularx}{\linewidth}{@{}>{\bfseries}lX@{}}

System: &
You are an audio understanding model that answers multiple-choice
questions based on audio content. \\[0.6em]

User: &
\texttt{<audio>}
Based on the audio between
\texttt{\{start\_time\}} and \texttt{\{end\_time\}},
\texttt{\{question\}}.
Please choose the answer from the following options:
\texttt{A. \{option\_A\}; B. \{option\_B\}; C. \{option\_C\}; D. \{option\_D\}.}
Output the final answer in
\texttt{<answer> </answer>}. \\[0.6em]

Assistant: &
\texttt{<answer>\{label\}</answer>}

\end{tabularx}
\end{tcolorbox}

\section{Conclusions}
We presented a segment-evidence-aware system for multilingual conversational Audio
MCQ. Timestamped ASR and LLM event extraction localize coherent evidence; dual
semantic and acoustic generators expand supervision; and quality plus trainability
verification separates invalid, easy, learnable, and currently unsolved candidates.
Verified weak items feed SFT, while audio-dependent items are optimized with a
debiased, importance-corrected, and dynamically filtered GSPO objective.
The final system reaches 90.92\% accuracy on the final evaluation set and
ranks 2nd. A limitation is that LLM-based event and
quality judgments may introduce correlated errors; future work will combine
independent verifiers and refresh trainability labels as the policy improves.

\section{Acknowledgements}
We thank the MLC-SLM Challenge organizers for providing the data and baselines.
We are also grateful to CAKE by VPBank for supporting this work, and to the DS/AI team at
CAKE for their valuable discussions, engineering support, and computational
resources.

\bibliographystyle{IEEEtran}
\bibliography{ref}

\end{document}